\documentclass[10pt,twocolumn]{article}

\usepackage{marvosym}
\usepackage{newtxtext,newtxmath}
\usepackage{microtype}

\usepackage[a4paper,left=1.5cm,right=1.5cm,top=2.2cm,bottom=2.2cm,columnsep=0.7cm]{geometry}

\usepackage[utf8]{inputenc}
\usepackage{amsmath}

\usepackage{amssymb}
\usepackage{graphicx}
\usepackage{booktabs}
\usepackage[table]{xcolor}
\usepackage{tikz}
\usetikzlibrary{arrows.meta,positioning,calc,decorations.pathreplacing}

\usepackage[numbers,sort&compress,super]{natbib}

\usepackage{caption}
\DeclareCaptionLabelFormat{nmi}{\textbf{Fig. #2}}
\DeclareCaptionLabelSeparator{nmi}{\textbf{ $|$ }}
\usepackage{titlesec}
\titleformat{\section}{\large\bfseries}{}{0em}{}
\titleformat{\subsection}{\normalsize\bfseries}{}{0em}{}
\titlespacing{\section}{0pt}{12pt plus 2pt minus 2pt}{4pt plus 1pt}
\titlespacing{\subsection}{0pt}{8pt plus 2pt minus 2pt}{3pt plus 1pt}

\usepackage{fancyhdr}
\fancypagestyle{firstpage}{%
  \fancyhf{}%
  \fancyfoot[R]{\footnotesize\thepage}%
}

\usepackage[hyphens]{url}
\usepackage[colorlinks=true,linkcolor=blue!70!black,citecolor=blue!70!black,urlcolor=blue!70!black,breaklinks=true]{hyperref}
\makeatletter
\g@addto@macro{\UrlBreaks}{\UrlOrds}
\makeatother

\begin{document}
\thispagestyle{firstpage}

\twocolumn[
\vspace*{-0.5cm}
{\noindent\small\textbf{Preprint}\par}
\vspace{12pt}

{\noindent\fontsize{22}{26}\selectfont\bfseries A Theory of LLM Information Susceptibility\par}
\vspace{14pt}

{\noindent\large Zhuo-Yang Song\textsuperscript{1,\,\Letter} \quad Hua Xing Zhu\textsuperscript{1,2,\,\Letter}\par}
\vspace{6pt}

{\noindent\footnotesize \textsuperscript{1}School of Physics, Peking University, Beijing 100871, China\\[2pt]
\textsuperscript{2}Center for High Energy Physics, Peking University, Beijing 100871, China\\[4pt]
{\Letter}\,e-mail: zhuoyangsong@stu.pku.edu.cn; zhuhx@pku.edu.cn\par}

\vspace{10pt}
\noindent\rule{\textwidth}{0.4pt}
\vspace{8pt}

{\small\noindent
Large language models (LLMs) are increasingly deployed as optimization modules in agentic systems, yet the fundamental limits of such LLM-mediated improvement remain poorly understood. Here we propose a theory of LLM information susceptibility, centred on the hypothesis that when computational resources are sufficiently large, the intervention of a fixed LLM does not increase the performance susceptibility of a strategy set with respect to budget. We develop a multi-variable utility-function framework that generalizes this hypothesis to architectures with multiple co-varying budget channels, and discuss the conditions under which co-scaling can exceed the susceptibility bound. We validate the theory empirically across structurally diverse domains and model scales spanning an order of magnitude, and show that nested, co-scaling architectures open response channels unavailable to fixed configurations. These results clarify when LLM intervention helps and when it does not, demonstrating that tools from statistical physics can provide predictive constraints for the design of AI systems. If the susceptibility hypothesis holds generally, the theory suggests that nested architectures may be a necessary structural condition for open-ended agentic self-improvement.
\par}
\vspace{8pt}
\noindent\rule{\textwidth}{0.4pt}
\vspace{16pt}
]

% ============================================================
%  INTRODUCTION
% ============================================================
\section*{Introduction}

Large language models (LLMs) are rapidly becoming core components of agentic systems, especially when combined with search, planning, verification, memory and tool-use modules~\cite{wang2023surveyautonomousagents,xi2023risepotentialagents,yao2023reactsynergizingreasoningacting,schick2023toolformer,Park2023,bran2023chemcrow,wang2023voyageropenendedembodiedagent,durante2024agentaisurveyingfoundations}. Such systems often outperform pure language modeling or traditional pipelines alone, motivating growing interest in agents that iteratively improve their own strategies, modules or internal organization~\cite{madaan2023selfrefine,shinn2023reflexion,zelikman2022star,funsearch,alphaevolve,EoH,song2025iteratedagentsymbolicregression}. Meanwhile, the empirical success of LLM-mediated optimization has outpaced our theoretical understanding of its limits. Existing work has focused primarily on specific prompting, training or inference schemes~\cite{wei2022chainofthought,wang2022selfconsistency,ouyang2022traininglanguagemodels,yao2023treeofthoughts,wang2023planandsolvepromptingimprovingzeroshot,chen2023programthoughtspromptingdisentangling,pmlr-v202-gao23f,Besta_Blach_Kubicek_Gerstenberger_Podstawski_Gianinazzi_Gajda_Lehmann_Niewiadomski_Nyczyk_Hoefler_2024,openai2024o1systemcard,deepseek2025r1}, but a general theoretical framework for understanding the fundamental limits of LLM-mediated optimization remains absent.

Here we propose a hypothesis about the limits of LLM-mediated optimization and, drawing on linear response theory~\cite{kubo1957statistical,de2017linear}, develop a framework to understand its applicability across different agent architectures. We treat an agent as producing a strategy set together with a utility function $J$ defined over that set, and study how $J$ changes with respect to computational variables that the architecture can control. This viewpoint is inherently broad: depending on the task and agent structure, $J$ may denote score, accuracy, ranking quality or another operational measure of performance, while the relevant budget variable $\mathcal{B}$ may denote beam width, search depth, sample count, model size, verification effort or other architecture-dependent resources. Within this formulation, we hypothesize that a fixed LLM-derived mapping cannot increase the performance susceptibility of the strategy set with respect to budget.

When there is only a single budget variable, this hypothesis can be equivalently expressed as a relative sensitivity $\alpha$ that has an upper bound of one in the large-budget regime. This hypothesis is significant because it separates two questions often conflated in discussions of agentic improvement: whether LLMs help in finite-budget settings (empirically, often yes~\cite{madaan2023selfrefine,shinn2023reflexion,wang2022selfconsistency,yao2023treeofthoughts}) and whether a fixed LLM layer can improve the asymptotic response of performance to additional computation. Our experiments address the latter question and give a negative answer in the fixed-architecture setting.

This provides a more precise way to reason about the design and optimization of high-compute pipelines and self-evolving agents~\cite{schick2023toolformer,Kaplan2020ScalingLaws,hoffmann2022training,kim2025sciencescalingagentsystems,song2025detailedbalancelargelanguage}. A self-evolving agent cannot simply be understood as a system that repeatedly applies the same optimization layer to its own outputs; rather, it must be a system whose performance-relevant components and computational channels change as complexity grows~\cite{shinn2023reflexion,wang2023voyageropenendedembodiedagent,funsearch,alphaevolve,song2025iteratedagentsymbolicregression}. We argue that, if the susceptibility hypothesis holds generally, nested architectures may be a necessary structural condition for overcoming the susceptibility bound imposed by a fixed LLM layer. More broadly, the framework developed here demonstrates that theoretical tools from statistical physics can provide a priori constraints and predictive structure in the design of complex agentic systems~\cite{de2017linear,Kaplan2020ScalingLaws,hoffmann2022training,wolpert2002no,kim2025sciencescalingagentsystems,song2025detailedbalancelargelanguage}.

% ============================================================
%  RESULTS
% ============================================================
\section*{Results}

\subsection*{A theory of LLM information susceptibility}

\begin{figure*}[!t]
    \centering
    \includegraphics[width=\textwidth]{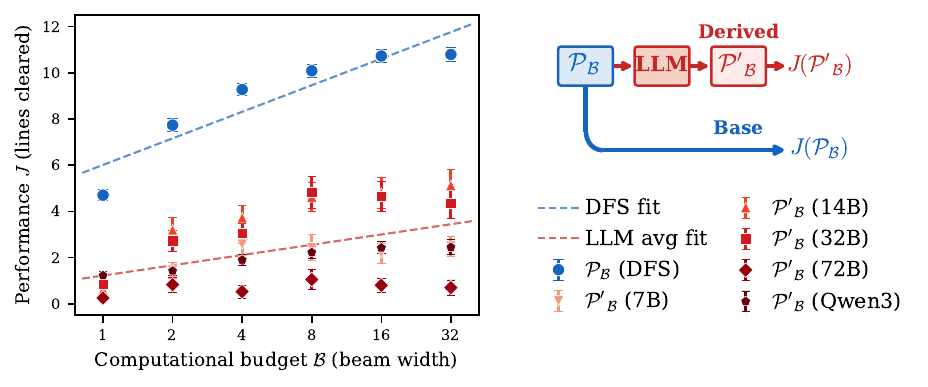}
    \caption{\textbf{Framework and representative results.} Performance $J$ (lines cleared) versus computational budget $\mathcal{B}$ (beam width) in the Tetris domain for the base algorithm (DFS, blue circles) and LLM-derived strategies (red markers; five Qwen models: 7B, 14B, 32B, 72B and Qwen3-Max). Dashed lines show linear fits for DFS and the LLM average. Error bars indicate the standard error of the mean across 40 random seeds. The schematic on the right illustrates the two evaluation paths: the base strategy set $\mathcal{P}_\mathcal{B}$ is evaluated directly by the utility function $J$ (base path, blue), or first processed by a fixed LLM to produce a derived set $\mathcal{P}'_\mathcal{B}$ (derived path, red).}
    \label{fig:tetris}
\end{figure*}

Consider a base strategy set $\mathcal{P}_\mathcal{B}$ generated under a computational budget $\mathcal{B}$ to maximize a utility function $J(\mathcal{P}_\mathcal{B})$ (see Fig.~\ref{fig:tetris}, right). As the computational resources increase without bound, $J(\mathcal{P}_{\mathcal{B}\to \infty})$ approaches its optimal value $J_{\infty}$. Now introduce a fixed LLM that reads the base strategy set $\mathcal{P}_\mathcal{B}$ and outputs a derived strategy set $\mathcal{P}'_\mathcal{B}$. We hypothesize that, when computational resources are sufficiently large, the performance susceptibility of $\mathcal{P}'_\mathcal{B}$ does not exceed that of $\mathcal{P}_\mathcal{B}$:
\begin{equation}
    \lim_{\mathcal{B}\to\infty}{\left\langle\frac{\partial J(\mathcal{P}_\mathcal{B})}{\partial \mathcal{B}}\right\rangle} \geq \lim_{\mathcal{B}\to\infty}{\left\langle\frac{\partial J(\mathcal{P}'_\mathcal{B})}{\partial \mathcal{B}}\right\rangle},
    \label{eq:principle}
\end{equation}
where $\langle\cdot\rangle$ denotes the average over different random seeds or experimental repetitions. This is the central hypothesis of the theory: the susceptibility $\partial J / \partial \mathcal{B}$ under the LLM-derived strategy cannot exceed that under the base strategy in the large-budget limit. The use of partial derivatives is deliberate: $J$ may in general depend on multiple budget variables, and this formulation provides the basis for the multi-variable generalization developed below. As a hypothesis, equation~(\ref{eq:principle}) is empirically testable and carries concrete implications for agent design: it implies that fixed LLM layers cannot improve the asymptotic scaling trajectory of the base strategy. Importantly, the asymptotic regime sets in at practically relevant budget levels: as shown in Fig.~\ref{fig:alpha}, the relative sensitivity $\alpha$ defined in equation~(\ref{eq:responsiveness}) crosses below 1 at $k \sim 12$ independent samples, after which the susceptibility bound is already operative. This rapid onset means the bound is not merely a theoretical limit but a constraint that governs real-world agent performance.

The intuition behind this claim rests on two arguments. First, as $\mathcal{B}\to\infty$, the performance $J(\mathcal{P}_\mathcal{B})$ converges toward the global optimum $J_\infty$, so the residual improvable gap $J_\infty - J(\mathcal{P}_\mathcal{B})$ shrinks. Any mapping applied to $\mathcal{P}_\mathcal{B}$, including the LLM, can only redistribute probability mass among strategies already present in or reachable from $\mathcal{P}_\mathcal{B}$; it cannot inject strategies that are not computable from the information contained in $\mathcal{P}_\mathcal{B}$ and the LLM's fixed parameters. Second, a fixed LLM can be viewed as a deterministic (or fixed-distribution) channel with finite capacity~\cite{shannon1948mathematical}: it compresses the input strategy set through a fixed-dimensional representation, based on its context window and parameters, and outputs a derived set. When the base set already encodes near-optimal information at large $\mathcal{B}$, the channel cannot amplify the marginal information content of additional budget. Since the mutual information between the derived set and the optimal strategy cannot exceed that between the base set and the optimal strategy by data-processing-inequality reasoning~\cite{cover2006elements}, the marginal return on budget cannot increase through the LLM intervention. This argument is not a formal proof, but it motivates why the bound $\alpha \leq 1$ should hold generically rather than being an artefact of specific tasks.

Figure~\ref{fig:tetris} shows representative results for the Tetris domain (see Methods for full experimental details). The performance of the base strategy set (beam search with depth-first backtracking, hereafter DFS) increases monotonically with beam width, while the LLM-derived strategy set exhibits a consistently lower susceptibility across all five Qwen-series models ranging from 7B to $\sim200$B parameters. A linear fit yields an average slope of 1.4 for the base algorithm versus 0.5 for the LLM-derived strategies, indicating that the LLM transforms each unit increase in beam width into about one third the performance gain of the base algorithm. This pattern is remarkably consistent: all five models, despite their order-of-magnitude difference in parameter count, fall within the same narrow performance band at each beam width, suggesting that the susceptibility bound is not merely a consequence of insufficient model capacity but reflects a structural property of the fixed-LLM intervention. We define the normalized performance gap as $\Delta(\mathcal{B}) = \left(J(\mathcal{P}_\mathcal{B}) - J(\mathcal{P}'_\mathcal{B})\right)/\overline{J(\mathcal{P}_\mathcal{B})}$, where $\overline{J(\mathcal{P}_\mathcal{B})}$ is the mean of the base performance over budget levels. The per-model breakdown of $\Delta(\mathcal{B})$ across all four domains is shown in Extended Data Fig.~\ref{fig:gap} and Extended Data Fig.~\ref{fig:gap_models}, confirming that this pattern holds at the level of individual models.

When the utility function $J$ depends on a single budget variable $\mathcal{B}$, the hypothesis can equivalently be expressed in terms of a relative sensitivity:
\begin{equation}
    \alpha(\mathcal{B}) = \left\langle\frac{d J(\mathcal{P}'_\mathcal{B})}{d J(\mathcal{P}_\mathcal{B})}\right\rangle = \frac{\left\langle \partial J(\mathcal{P}'_\mathcal{B})/\partial \mathcal{B}\right\rangle}{\left\langle \partial J(\mathcal{P}_\mathcal{B})/\partial \mathcal{B}\right\rangle} \leq 1 \quad (\mathcal{B}\to\infty).
    \label{eq:responsiveness}
\end{equation}
Here $\alpha(\mathcal{B})$ admits a natural interpretation: computational resources increase the mutual information between the strategy set and the optimum, while the fixed LLM channel cannot amplify this information gain (by the data-processing inequality), so that $\alpha \leq 1$ when computational resources are sufficiently large.

\subsection*{Robustness of the susceptibility bound}

\begin{figure*}[!t]
    \centering
    \includegraphics[width=\textwidth]{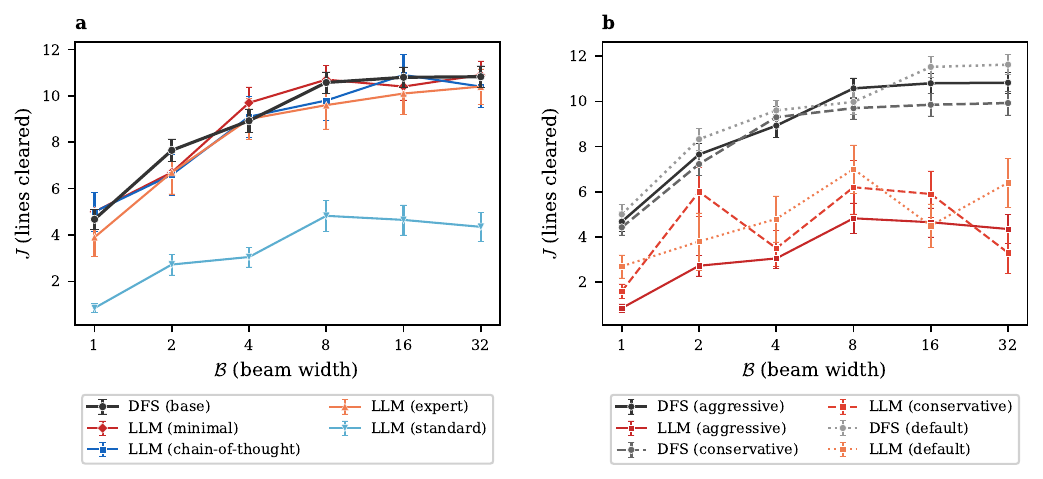}
    \caption{\textbf{Robustness of the susceptibility bound.} \textbf{a}, Four prompt variants compared against the DFS baseline in the Tetris domain (Qwen-32B). The minimal prompt nearly matches DFS at high $\mathcal{B}$, while more elaborative prompts amplify the gap. \textbf{b}, Three reward functions overlaid for both DFS (grey shades) and LLM (red shades, Qwen-32B). The susceptibility bound holds across all prompt and reward configurations.}
    \label{fig:robustness}
\end{figure*}

A natural concern is whether the observed susceptibility gap is an artefact of specific prompt engineering choices or reward function design. We tested both systematically in the Tetris domain (Fig.~\ref{fig:robustness}). Four prompt variants were evaluated: minimal (JSON-only output), standard (full analysis), chain-of-thought (5-step reasoning) and expert (domain-specific strategy). All variants exhibit the same qualitative behaviour: the susceptibility of the LLM-derived strategy does not exceed that of the base strategy (Fig.~\ref{fig:robustness}a). The observation that the minimal prompt, which provides the least guidance to the LLM, nearly matches the DFS baseline implies that the gap arises from active reprocessing rather than a passive information bottleneck. Three distinct reward functions likewise show qualitative invariance: in all cases the DFS baseline outperforms the LLM-derived strategy and the gap grows with budget (Fig.~\ref{fig:robustness}b), confirming that the susceptibility bound is a structural property of the fixed-LLM intervention, independent of prompt design or reward signal.

\subsection*{Empirical characterization of the sufficiency condition}

The theory predicts that the susceptibility bound holds when computational resources are ``sufficiently large'', but does not specify the threshold a priori. To characterize this transition empirically, we designed an experiment using 60 mathematics problems from AIME 2024 and 2025~\cite{hendrycks2021measuringmathematicalproblemsolving,huggingface2024aime}. In this domain the performance depends on three variables: $J = J(k, \mathcal{B}_\mathrm{gen}, \mathcal{B}_\mathrm{sel})$, where $k$ is the number of independent solution attempts, $\mathcal{B}_\mathrm{gen}$ is the generator model size and $\mathcal{B}_\mathrm{sel}$ is the selector model size. A generator LLM of size $\mathcal{B}_\mathrm{gen}$ produces $k$ independent solution attempts, and the base strategy applies majority vote~\cite{wang2022selfconsistency}. A fixed selector LLM of size $\mathcal{B}_\mathrm{sel}$ then reads the candidate answers and selects one, forming the derived strategy set $\mathcal{P}'_\mathcal{B}$. This generate-then-select architecture has been widely adopted in competitive programming~\cite{li2022alphacode}, mathematical reasoning~\cite{cobbe2021training,brown2024large} and scientific discovery~\cite{funsearch,alphaevolve}.

To isolate the effect of the sample budget $k$, we average over all five selector model sizes $\mathcal{B}_\mathrm{sel}$ and all five generator model sizes $\mathcal{B}_\mathrm{gen}$. This yields an average sensitivity $\bar{\alpha}(k) = \langle \alpha(\mathcal{B}_\mathrm{gen}, \mathcal{B}_\mathrm{sel}; k) \rangle_{\mathcal{B}_\mathrm{gen}, \mathcal{B}_\mathrm{sel}}$ that characterizes how the relative advantage of the LLM selector evolves as the base strategy aggregates more samples.

\begin{figure}[!t]
    \centering
    \includegraphics[width=\columnwidth]{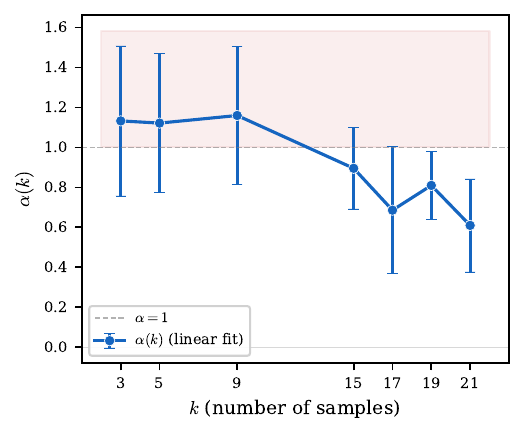}
    \caption{\textbf{Transition of the relative sensitivity $\alpha$.} Average $\alpha$ versus the number of samples $k$. For each $k$, $\alpha$ is estimated by fitting $J_\mathrm{agent} = \alpha \cdot J_\mathrm{MV} + \beta$ across five generator model sizes, where $J_\mathrm{MV}$ is the majority-vote accuracy, $J_\mathrm{agent}$ is the LLM-selector accuracy and $\beta$ is the regression intercept (see Methods). As $k$ increases, $\alpha$ decreases and falls below 1 around $k \sim 12$, marking the onset of the large-budget regime where the susceptibility bound takes effect.}
    \label{fig:alpha}
\end{figure}

Figure~\ref{fig:alpha} shows $\bar{\alpha}(k)$ as a function of $k$. At low $k$ ($\leq 5$), the selector LLM could outperform majority vote ($\bar{\alpha} > 1$), reflecting the regime in which the LLM's world knowledge and reasoning provide a genuine advantage over a sparse vote distribution. As $k$ increases, $\bar{\alpha}$ crosses below 1 and continues to decline, marking the onset of the large-budget regime in which majority vote becomes statistically robust and the fixed selector can no longer improve upon it. This crossover~\cite{HOGG19961} provides an empirical operationalization of ``sufficiently large'': where the base strategy's aggregation of diverse samples begins to dominate the LLM's judgement.

\subsection*{Cross-domain validation}

\begin{figure*}[!t]
    \centering
    \includegraphics[width=\textwidth]{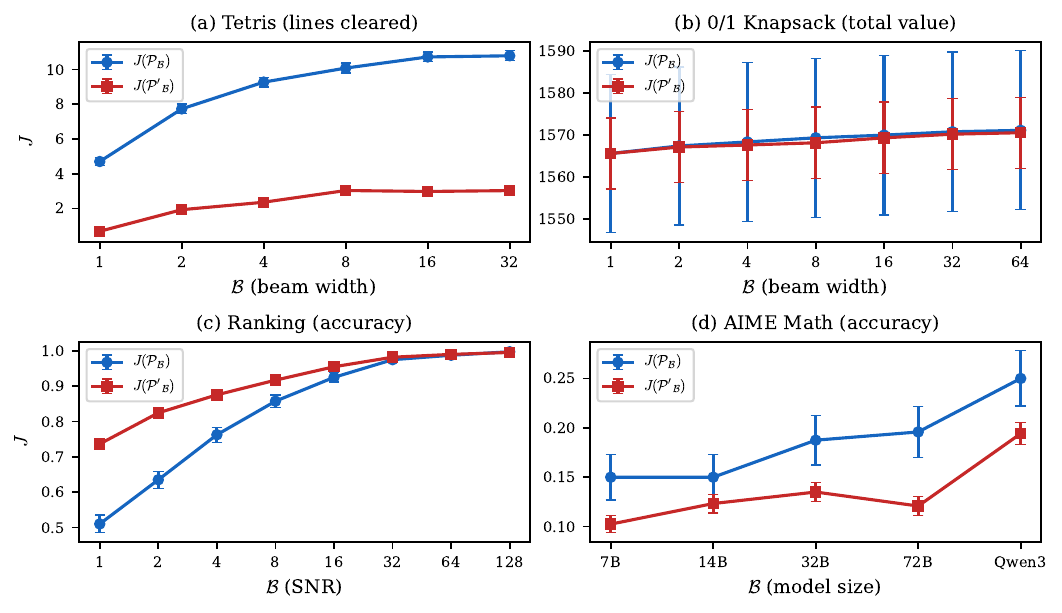}
    \caption{\textbf{Cross-domain validation.} Performance $J$ versus computational budget $\mathcal{B}$ for the base strategy set (blue circles) and the LLM-derived strategy set (red squares) across four domains: Tetris, Knapsack, Ranking and AIME mathematics. In the AIME domain, the derived strategy set averages over all five selector models and over $k \in \{15, 17, 19, 21\}$.}
    \label{fig:crossdomain}
\end{figure*}

To test the universality of the hypothesis, we conducted experiments across four task domains that differ substantially in their structure and the role of LLM knowledge: Tetris (combinatorial game-playing), 0/1 Knapsack~\cite{kellerer2004knapsack} (combinatorial optimization), world-knowledge Ranking (factual recall under noise) and AIME mathematics (multi-step reasoning). Full experimental configurations are provided in Methods; results are shown in Fig.~\ref{fig:crossdomain}.

Across all domains, the base strategy set's performance increases monotonically with computational budget, while the LLM-derived strategy set's susceptibility is generally not larger, validating equation~(\ref{eq:principle}). The Ranking domain is particularly instructive: at low budgets, the LLM significantly outperforms the noisy algorithmic baseline because it can draw on world knowledge (for example, identifying China as more populous than Japan regardless of the noisy score estimate). However, as the signal-to-noise ratio increases, the algorithmic ranking converges to the ground truth and the LLM advantage vanishes, consistent with the hypothesis's prediction that the susceptibility advantage of the base strategy dominates at large budget. This pattern, greater LLM advantage at low budget and greater algorithmic advantage at high budget, is precisely the signature predicted by the theory and is observed across all four domains.

These results underscore that the utility function $J$ is flexible and task-dependent: it may represent game score, solution quality, ranking performance or answer accuracy, depending on the domain. Likewise, $\mathcal{B}$ should be interpreted as the controllable budget associated with the underlying agent.
The Knapsack domain deserves particular comment: the performance gap $\Delta(\mathcal{B})$ is nearly zero across all budget levels and model sizes (Extended Data Fig.~\ref{fig:gap}). This is consistent with the theory ($\alpha \leq 1$) but does not exhibit the dramatic separation seen in Tetris. The likely explanation is that the LLM acts approximately as an identity mapping in this domain: because the beam-search candidates are already sorted by value density and the packing structure is opaque to the LLM without explicit combinatorial reasoning, the model largely defers to the algorithmic ranking rather than reprocessing it. This ``pass-through'' regime is similar to the minimal prompt regime in Tetris and represents a qualitatively different manifestation of the susceptibility bound, one in which the LLM neither helps nor hurts, because it recognizes the limits of its own intervention. The empirical evidence therefore supports a general statement: the hypothesis applies whenever one can define a strategy set, a utility function over that set and a meaningful computational variable with respect to which susceptibility is measured.

\subsection*{Generality: $J$ as a multi-variable utility function}

The experiments also clarify the scope of the framework. The basic formulation in equation~(\ref{eq:principle}) concerns a fixed derivation mapping responsive to a single effective budget variable. More generally, the utility function $J$ depends on all architectural budget variables: $J = J(\mathcal{B}_1, \mathcal{B}_2, \ldots, \mathcal{B}_n)$. By analogy with linear response theory~\cite{kubo1957statistical}, the gradient $\nabla_{\mathcal{B}} J$ is the susceptibility vector; each component $\partial J / \partial \mathcal{B}_i$ measures how efficiently one budget channel converts additional compute into performance. Equation~(\ref{eq:responsiveness}) is the $n = 1$ special case in which a single budget variable controls the entire system.

When the architecture is extended so that additional computational variables become relevant, the utility function can be correspondingly generalized. If we write $J$ for the seed-averaged derived-strategy performance and $J_\mathrm{base}$ for that of the base strategy, both as deterministic functions of the budget variables, the generalized total sensitivity follows by summing over all budget channels that co-vary with a reference budget $\mathcal{B}_\mathrm{ref}$:
\begin{equation}
    \alpha_\mathrm{total} = \sum_{i=1}^{n} \frac{\partial J / \partial \mathcal{B}_i}{\partial J_\mathrm{base} / \partial \mathcal{B}_\mathrm{ref}} \cdot \frac{d \mathcal{B}_i}{d \mathcal{B}_\mathrm{ref}}.
    \label{eq:total_susceptibility}
\end{equation}
As a concrete example, in the AIME domain, if the selector LLM is allowed to vary with the generator LLM, then the utility of the derived strategy set becomes $J(\mathcal{P}'_{\mathcal{B}_\mathrm{gen}}, \mathcal{B}_\mathrm{sel})$. Here the results are averaged over large values of $k$ ($k \in \{15, 17, 19, 21\}$), which is therefore not treated as a co-varying budget channel. Setting $n = 2$, $\mathcal{B}_1 = \mathcal{B}_\mathrm{gen}$, $\mathcal{B}_2 = \mathcal{B}_\mathrm{sel}$ and $\mathcal{B}_\mathrm{ref} = \mathcal{B}_\mathrm{gen}$, the relative sensitivity reduces to
\begin{align}
    \alpha(\mathcal{B}_\mathrm{gen}, \mathcal{B}_\mathrm{sel})
    &= \frac{\partial J(\mathcal{P}'_{\mathcal{B}_\mathrm{gen}}, \mathcal{B}_\mathrm{sel})/\partial \mathcal{B}_\mathrm{gen}}{\partial J(\mathcal{P}_{\mathcal{B}_\mathrm{gen}})/\partial \mathcal{B}_\mathrm{gen}} \nonumber \\
    &\quad + \frac{\partial J(\mathcal{P}'_{\mathcal{B}_\mathrm{gen}}, \mathcal{B}_\mathrm{sel})/\partial \mathcal{B}_\mathrm{sel}}{\partial J(\mathcal{P}_{\mathcal{B}_\mathrm{gen}})/\partial \mathcal{B}_\mathrm{gen}} \cdot \frac{d \mathcal{B}_\mathrm{sel}}{d \mathcal{B}_\mathrm{gen}}.
    \label{eq:nested}
\end{align}
The first term on the right-hand side is the fixed-architecture contribution constrained by the hypothesis ($\alpha \leq 1$), while the second term appears only when the architecture itself is allowed to vary with budget. Here $d\mathcal{B}_\mathrm{sel}/d\mathcal{B}_\mathrm{gen}$ is the rate at which the selector's budget changes when the generator's budget is increased: it equals zero in the fixed-selector configuration and one when generator and selector are co-scaled.

Note that equation~(\ref{eq:total_susceptibility}) has a covariant-contravariant structure: the susceptibility vector $\partial J / \partial \mathcal{B}_i$ characterizes the local geometry of the performance landscape (partial derivatives hold other budget variables fixed), while the scaling protocol $d\mathcal{B}_i / d\mathcal{B}_\mathrm{ref}$ is a design choice specifying how budget channels co-vary. Their contraction yields the scalar $\alpha_\mathrm{total}$, which depends on both the landscape and the chosen scaling path.

\begin{figure*}[!t]
\centering
\includegraphics[width=\textwidth]{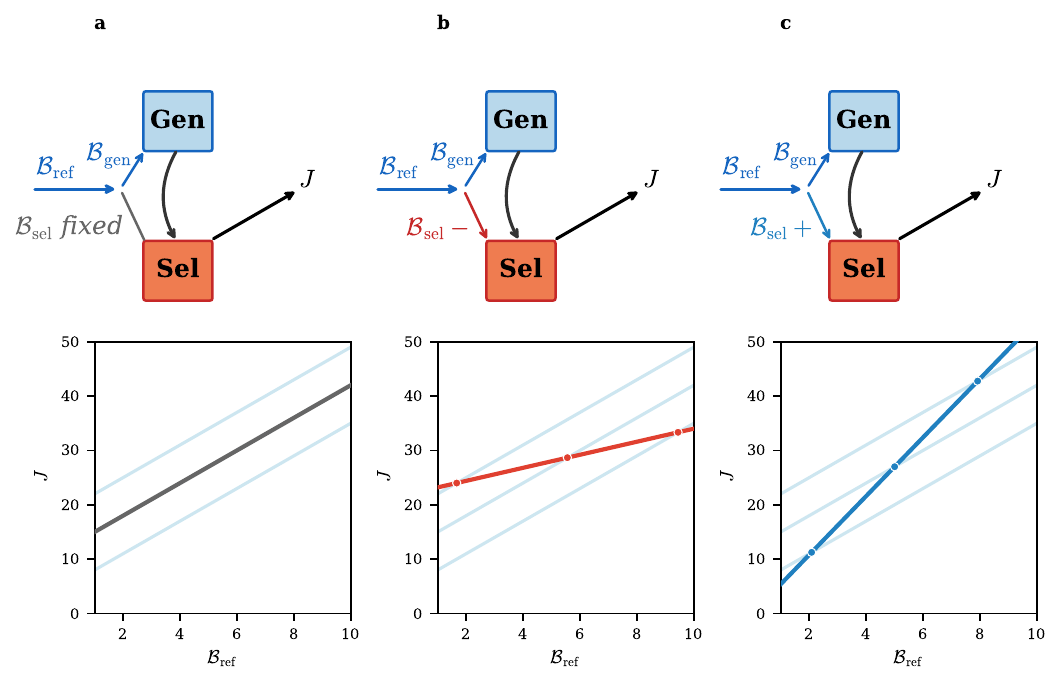}
\caption{\textbf{Inter-layer coupling regimes.} Each panel shows an architecture diagram (top) and illustration of $J$ versus $\mathcal{B}_\mathrm{ref}$ (bottom). Faded blue lines represent three fixed-selector configurations; solid coloured lines show the nested (co-scaled) configuration. Dots mark intersection points where configurations coincide. \textbf{a}, Decoupled: only the generator scales with $\mathcal{B}_\mathrm{ref}$; the selector remains fixed. The nested line coincides with one of the fixed lines. \textbf{b}, Negative coupling: both components scale, but co-scaling reduces marginal return ($\alpha_\mathrm{total} < 1$). The nested line falls below the fixed line. \textbf{c}, Positive coupling: co-scaling amplifies marginal return ($\alpha_\mathrm{total}$ can exceed 1). The nested line exceeds all fixed lines, opening a response channel unavailable to fixed architectures.}
\label{fig:response}
\end{figure*}

\begin{figure*}[!t]
    \centering
    \includegraphics[width=0.65\textwidth]{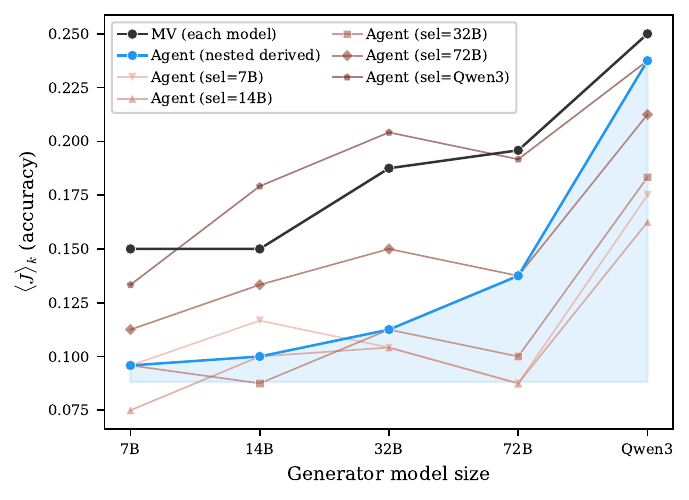}
    \caption{\textbf{Nested versus fixed architectures in the AIME domain.} Accuracy $J$ (averaged over $k \in \{15, 17, 19, 21\}$) versus model size for the nested derived strategy set (generator and selector co-scaled) and fixed derived strategy sets (fixed selector, varying generator). The curves intersect at the model size of the respective fixed selector, showing that co-scaling architectural components can exceed the susceptibility bound.}
    \label{fig:nested}
\end{figure*}

This viewpoint reveals three distinct coupling regimes (Fig.~\ref{fig:response}). (i)~\emph{Decoupled} ($d\mathcal{B}_\mathrm{sel}/d\mathcal{B}_\mathrm{ref} = 0$): each budget channel operates independently, and the hypothesis $\alpha \leq 1$ applies to each channel separately; this is the regime described by equation~(\ref{eq:responsiveness}). (ii)~\emph{Negative coupling}: In this regime, co-scaling the selector with the generator reduces the marginal return of additional budget, analogous to Le Chatelier's principle~\cite{lechatelier1884}, so that the total slope $\alpha_\mathrm{total} < \alpha_\mathrm{gen} \leq 1$ falls below that of the fixed-selector curve, where $\alpha_\mathrm{gen}$ denotes the first term on the right-hand side of equation~(\ref{eq:nested}), the contribution from the generator channel alone (Fig.~\ref{fig:response}b). This occurs when $\partial J / \partial \mathcal{B}_\mathrm{sel}$ and $d\mathcal{B}_\mathrm{sel}/d\mathcal{B}_\mathrm{ref}$ have opposite signs, so that their product contributes a negative term to $\alpha_\mathrm{total}$. (iii)~\emph{Positive coupling}: co-scaling increases the marginal return, so that $\alpha_\mathrm{total}$ can exceed~1 (Fig.~\ref{fig:response}c). This occurs when a stronger selector genuinely complements a stronger generator, as demonstrated empirically in the nested AIME configuration (equation~(\ref{eq:nested}) and Fig.~\ref{fig:nested}).

% --- Three-panel coupling schematic (Python-generated) ---

The sign of the inter-layer coupling can be estimated empirically from how the utility function changes with different budget combinations: positive coupling indicates that increasing the generator's capability amplifies the marginal return of the selector, and vice versa. When the coupling is positive, co-scaling is beneficial and a nested architecture is preferred; when it is near zero or negative, independent scaling of individual components may be more efficient. This provides a concrete, measurable design criterion: before committing to a nested agent architecture, evaluate $\alpha_\mathrm{total}$ from a small grid of budget combinations and check whether co-scaling improves the marginal return.

Figure~\ref{fig:nested} illustrates this in the AIME domain: we compare a ``nested'' configuration, in which the generator and selector are the same model and thus co-scale, against ``fixed'' configurations, in which the selector is held constant while the generator varies. The nested curve intersects each fixed-selector curve at the model size of the respective fixed selector, since the two configurations coincide at that point. Crucially, the nested curve can exceed any individual fixed-selector curve in the large-generator regime, demonstrating that co-scaling architectural components opens a response channel that is not available to the fixed-layer configuration. The fixed-architecture hypothesis applies to each individual fixed-selector curve, but does not constrain the nested curve, which can exceed the envelope of the fixed-selector family and thereby explore a fundamentally different region of the architectural parameter space.

% ============================================================
%  DISCUSSION
% ============================================================
\section*{Discussion}

A theory of LLM information susceptibility addresses a question that is increasingly pressing as LLM-based agents are deployed in high-compute settings: does inserting a fixed LLM layer into an optimization pipeline improve how efficiently additional computation is converted into performance? Our results give a negative answer for fixed architectures and a conditionally positive answer for nested, co-scaling ones.

This finding has a natural interpretation in terms of the susceptibility framework. The utility function $J$ is not defined independently of architecture: the structure of the agent determines which budget variables are available, how they couple to one another and which response channels contribute to performance~\cite{kubo1957statistical,de2017linear,Kaplan2020ScalingLaws,hoffmann2022training,snell2024scaling,kim2025sciencescalingagentsystems}. The generalized susceptibility (equation~(\ref{eq:total_susceptibility})) makes this dependence explicit: the contraction of the susceptibility vector $\partial J / \partial \mathcal{B}_i$ with the scaling protocol $d\mathcal{B}_i / d\mathcal{B}_\mathrm{ref}$ determines whether co-scaling helps or hurts (Fig.~\ref{fig:response}). If the agent structure is held fixed and only the budget along one response channel is increased, then LLM intervention can improve constants or finite-budget behaviour, but it does not increase the large-budget susceptibility. By contrast, nesting changes the relationship between $J$ and its budget variables by allowing the capability of one component to scale with the complexity induced by another, a regime characterized by positive inter-layer coupling. This perspective is consistent with the potential-landscape analysis of Song et al.~\cite{song2025detailedbalancelargelanguage}, which shows that within a fixed LLM-driven agent, optimization is constrained by an intrinsic landscape. Our results complement that picture at the system level: repeated optimization by a fixed layer is fundamentally limited both by internal model structure and by external response structure.

These findings carry practical implications for agent design. First, when the target application operates in a large-budget regime, investing computation in the base strategy-generation process, like stronger search, better proposal generation or more reliable verification, may be more effective than relying on a fixed LLM wrapper to amplify gains~\cite{HOGG19961,Kaplan2020ScalingLaws,hoffmann2022training,snell2024scaling,kim2025sciencescalingagentsystems}. Second, static LLM selection modules are most useful in low- or intermediate-budget regimes, where world knowledge and heuristic compression still provide noticeable improvements~\cite{madaan2023selfrefine,wang2022selfconsistency,shinn2023reflexion}. Third, if the goal is to build systems capable of open-ended improvement, designers should allocate budget so that generator, selector, verifier, memory and tool-use components can co-scale~\cite{schick2023toolformer,wang2023voyageropenendedembodiedagent,alphaevolve,kim2025sciencescalingagentsystems,hu2024automateddesignagenticsystems,song2025iteratedagentsymbolicregression,hosseini2024vstar}. More broadly, the susceptibility-based viewpoint developed here suggests a quantitative language for comparing agent architectures: rather than asking only whether an LLM helps, one can ask which architectural variables appear in $J$, how those variables couple through the scaling protocol and which susceptibilities dominate in the regime of interest~\cite{wolpert2002no,kim2025sciencescalingagentsystems}.

Beyond these design implications, the results bear directly on a fundamental question in AI: whether LLMs can achieve open-ended self-evolution~\cite{good1966speculations,zelikman2022star,singh2024beyond} (see Extended Data Fig.~\ref{fig:selfevolution} for a detailed phenomenological model). Consider a scenario in which an LLM attempts to improve its own strategies by using itself as the optimization layer. If the LLM mediation cannot increase asymptotic susceptibility ($\alpha \leq 1$), then self-guided improvement is expected to saturate once the model's capability exceeds a threshold, because the fixed LLM layer cannot increase the rate at which performance responds to additional computation; the feedback loop of self-improvement is inherently bounded. A related limitation has been observed in unsupervised reinforcement learning, where initial training gains are followed by collapse once the self-generated reward signal diverges from the true objective at sufficient scale~\cite{he2026urlvr}. Conversely, if a nested architecture enables $\alpha_\mathrm{total} > 1$, the LLM can alter its own strategy distribution in a way that increases marginal return: as the LLM's capability grows, its ability to guide its own improvement strengthens in turn, potentially creating a positive feedback loop. Figure~\ref{fig:nested} provides empirical evidence for this logic: the nested configuration's accuracy is approaching and poised to exceed the majority-vote baseline, indicating that the LLM's ability to reshape its own distribution through nested co-scaling is nearing a critical crossover. In the current experiments, the nested curve for Qwen3-Max is close to but has not yet crossed this threshold. Contingent on the susceptibility hypothesis holding, this suggests that nested, co-scaling architectures are not merely sufficient for exceeding the susceptibility bound, but constitute a necessary structural condition for open-ended self-evolution: if fixed architectures cannot achieve $\alpha > 1$, only architectures whose components co-scale can sustain unbounded improvement.

Several directions for future work emerge naturally from this study. First, the theory is stated as an empirical hypothesis supported by experiments; developing a formal proof would place the bound on firmer theoretical ground. Second, the four domains tested, though structurally diverse, do not cover settings with very long horizons, multi-agent interaction or continuous action spaces, where the relationship between budget and performance may differ qualitatively; exploring these settings would clarify the boundary conditions of the framework. Third, the framework suggests a practical engineering methodology: by measuring the susceptibility of individual architectural layers and combining these measurements with the known compositional structure of the architecture, one could in principle, if the inter-layer coupling structure is known, reconstruct the full utility function $J$ across the entire budget space; this would reduce system-level performance prediction from costly end-to-end evaluation to composable single-layer characterizations, offering more efficient guidance for engineering design. Finally, the nested-architecture experiments demonstrate that co-scaling can exceed the susceptibility bound, but do not yet characterize the rate at which it does so; deriving a quantitative scaling law for nested susceptibility is perhaps the most important open question, as it would provide concrete guidance for allocating compute across co-scaling components.

Beyond these future directions, the framework offers a concrete criterion for evaluating when LLM intervention is worth the cost: compute the sensitivity $\alpha$ in the target budget regime. If $\alpha < 1$, the LLM layer is consuming resources without proportionally improving the scaling trajectory, and the design should either move to a nested architecture or redirect computation to the base strategy. This criterion is measurable, domain-agnostic and complementary to standard metrics such as absolute accuracy or win rate that do not distinguish between constant offsets and scaling improvements. More generally, the susceptibility-based approach demonstrates that tools from statistical physics can provide a predictive framework for the study of AI systems, one that constrains design choices beyond post-hoc rationalization of empirical results. Among its concrete, hypothesis-dependent predictions is that open-ended self-evolution may require nested co-scaling, a claim that is already approaching testability with current models.

% ============================================================
%  METHODS
% ============================================================
\section*{Methods}

\subsection*{Models and infrastructure}

All experiments use five Qwen-series models: Qwen-2.5-7B-Instruct (7B), Qwen-2.5-14B-Instruct (14B), Qwen-2.5-32B-Instruct (32B), Qwen-2.5-72B-Instruct (72B) and Qwen3-Max (${\sim}200$B)~\cite{qwen2025qwen25technicalreport,yang2025qwen3technicalreport}. Decoding parameters are specified per domain below. All domains use the same models and API, ensuring that the observed effects are not artefacts of a particular model.

\subsection*{Tetris}

\textbf{Environment.} A $10 \times 20$ Tetris board with 6 pre-filled garbage lines. Pieces are drawn from 18 fixed orientations (I, O, T, S, Z, L, J variants); no rotation is performed during play. Each game lasts at most 50 steps. The utility function $J$ is the number of lines cleared.

\textbf{Base strategy $\mathcal{P}_\mathcal{B}$.} 
Beam search~\cite{russell2021artificial} with depth-first backtracking and a lookahead depth of 3. At each step, the algorithm expands all legal placements to depth 3, evaluates terminal states using a heuristic combining aggregate height, hole count, bumpiness and lines cleared, and retains the top-$\mathcal{B}$ candidates (beam width). The top 3 placements are returned as candidates. Beam widths tested: $\mathcal{B} \in \{1, 2, 4, 8, 16, 32\}$.

\textbf{Derived strategy $\mathcal{P}'_\mathcal{B}$.} Each LLM receives the current board state (ASCII grid), the current piece and the top 3 DFS candidates with their heuristic scores. The LLM selects one placement. Decoding: temperature $= 0.1$, max tokens $= 500$, timeout $= 15$\,s, max retries $= 2$.

\textbf{Prompt variants.} Four prompt designs were tested: minimal (JSON-only output format), standard (full board analysis), chain-of-thought (explicit 5-step reasoning) and expert (domain-specific Tetris strategy). The main text reports results using the standard prompt with the aggressive reward function as the representative case showing the strongest susceptibility gap; robustness across all prompt and reward configurations is reported in Fig.~\ref{fig:robustness}.

\textbf{Reward functions.} Three heuristic evaluation functions were tested: aggressive (prioritizing line clearing with weight 5.0), conservative (prioritizing hole avoidance with weight 3.0) and default (balanced weights). The qualitative pattern of the susceptibility bound is invariant across all three.

\textbf{Statistics.} 40 independent random seeds per (model, $\mathcal{B}$) pair. Error bars in Fig.~\ref{fig:tetris} are standard errors of the mean over seeds.

\subsection*{AIME mathematics}

\textbf{Problem set.} 60 problems from AIME 2024 (30 problems) and AIME 2025 (30 problems). Each answer is an integer in $[0, 999]$.

\textbf{Base strategy $\mathcal{P}_\mathcal{B}$.} For each problem, a generator LLM of size $\mathcal{B}_\mathrm{gen}$ produces $k$ independent solution attempts at temperature 0.7 (max tokens $= 1{,}500$). The base strategy applies majority vote~\cite{condorcet1785essai,wang2022selfconsistency}: answers are grouped by approximate equality ($|a - b| < 0.5$) and the most common group is selected, with random tie-breaking. Here $k$ serves as a control parameter that tunes the statistical power of the majority vote, while the generator model size $\mathcal{B}_\mathrm{gen}$ determines the quality of individual attempts. Values tested: $k \in \{1, 3, 5, 9, 15, 17, 19, 21\}$; all 21 samples are generated once and subsampled for each $k$.

\textbf{Derived strategy $\mathcal{P}'_\mathcal{B}$.} A selector LLM of size $\mathcal{B}_\mathrm{sel}$ reads the $k$ candidate answers (deduplicated, without frequency counts) and selects one. The ``fixed derived'' configuration uses each of the five models as a fixed selector while varying the generator model; the reported $\bar{\alpha}(k)$ averages over all five generator sizes and all five selectors. Agent selection uses temperature $= 0.1$. The generation temperature of 0.7 ensures diversity across the $k$ independent attempts, while the low selection temperature yields deterministic selector behaviour. The prompt does not strictly adhere to the official AIME format; this is intentional, to minimize wording differences between the majority-vote and LLM-selector conditions.

\textbf{Estimation of $\alpha$.} For each $k$ and each fixed selector, five data points $(J_\mathrm{MV}^{(i)}, J_\mathrm{agent}^{(i)})$ are obtained, one per generator model size $\mathcal{B}_\mathrm{gen}$. A linear model $J_\mathrm{agent} = \alpha \cdot J_\mathrm{MV} + \beta$ is fitted using ordinary least squares. The slope $\alpha$ and its standard error are reported. The average $\bar{\alpha}(k)$ shown in Fig.~\ref{fig:alpha} is obtained by first averaging the agent's accuracy over all five selectors for each generator size, then fitting a single linear model across the five generator sizes.

\textbf{Statistics.} The accuracy for each (model, $k$) pair is the mean correctness over 60 problems. Error bars in Figs.~\ref{fig:crossdomain} and~\ref{fig:nested} are binomial standard errors $\sqrt{p(1-p)/n}$, where $p$ is the observed accuracy and $n$ is the number of independent trials. For the majority-vote baseline, $n = 60 \times |K|$ (60 problems times the number of $k$ values averaged over); for the LLM agent, $n = 60 \times 5 \times |K|$ (additionally averaged over five selector configurations). The binomial standard error is used because each problem outcome is a Bernoulli trial (correct or incorrect), and the standard error quantifies the uncertainty due to finite sample size.

\subsection*{0/1 Knapsack}

50 items with weights $w_i \in [1, 50]$ and values $v_i \in [1, 100]$, capacity $= 0.3 \sum w_i$. The base strategy is beam search over the item-selection tree, with items sorted by value density $v_i/w_i$~\cite{kellerer2004knapsack}. The LLM receives the top 3 packings and selects one. $J$ = total value, $\mathcal{B}$ = beam width $\in \{1, 2, 4, 8, 16, 32, 64\}$. Statistics: 50 problem instances; error bars are standard errors of the mean over instances.

\subsection*{World-knowledge Ranking}

Four real-world ranking datasets (GDP of 15 countries, population of 15 countries, diameters of 8 planets, weights of 12 animals). For each item, a noisy score estimate is generated: $\hat{s}_i = s_i + \mathcal{N}(0, \sigma/\sqrt{\mathcal{B}})$, where $\sigma$ is a dataset-specific baseline noise scale chosen so that the algorithmic success rate is approximately 50\% at $\mathcal{B} = 1$. The top 5 candidates by noisy score are presented to the LLM, which selects the item it believes ranks first using world knowledge. $J$ = fraction correctly identifying the true rank-1 item, $\mathcal{B}$ = signal-to-noise ratio $\in \{1, 2, 4, 8, 16, 32, 64, 128\}$. Statistics: 100 noise seeds $\times$ 4 datasets $\times$ 8 SNR levels. Error bars are standard errors of the mean over noise seeds and datasets.

% ============================================================
%  DATA AND CODE AVAILABILITY
% ============================================================
\section*{Data availability}
All experimental data generated in this study are publicly available on HuggingFace at \url{https://huggingface.co/datasets/Nondegeneracy/LLM-Susceptibility-theory} under the CC BY 4.0 license.

\section*{Code availability}
The code used to run the experiments and produce all figures is available on GitHub at \url{https://github.com/SonnyNondegeneracy/LLM-Susceptibility-theory} under the MIT license.

% ============================================================
%  ACKNOWLEDGEMENTS
% ============================================================
\section*{Acknowledgements}
This work is supported by National Natural Science Foundation of China under contract No. 12425505.

% ============================================================
%  COMPETING INTERESTS
% ============================================================
\section*{Competing interests}
The author declares no competing interests.

% ============================================================
%  REFERENCES
% ============================================================
\bibliography{cite}

@misc{song2025detailedbalancelargelanguage,
      title={Detailed balance in large language model-driven agents}, 
      author={Zhuo-Yang Song and Qing-Hong Cao and Ming-xing Luo and Hua Xing Zhu},
      year={2025},
      eprint={2512.10047},
      archivePrefix={arXiv},
      primaryClass={cs.LG},
      url={https://arxiv.org/abs/2512.10047}, 
}

@misc{hendrycks2021measuringmathematicalproblemsolving,
      title={Measuring Mathematical Problem Solving With the MATH Dataset}, 
      author={Dan Hendrycks and Collin Burns and Saurav Kadavath and Akul Arora and Steven Basart and Eric Tang and Dawn Song and Jacob Steinhardt},
      year={2021},
      eprint={2103.03874},
      archivePrefix={arXiv},
      primaryClass={cs.LG},
      url={https://arxiv.org/abs/2103.03874}, 
}

@misc{song2025iteratedagentsymbolicregression,
      title={Iterated Agent for Symbolic Regression}, 
      author={Zhuo-Yang Song and Zeyu Cai and Shutao Zhang and Jiashen Wei and Jichen Pan and Shi Qiu and Qing-Hong Cao and Tie-Jiun Hou and Xiaohui Liu and Ming-xing Luo and Hua Xing Zhu},
      year={2025},
      eprint={2510.08317},
      archivePrefix={arXiv},
      primaryClass={physics.comp-ph},
      url={https://arxiv.org/abs/2510.08317}, 
}

@article{wolpert2002no,
  title={No free lunch theorems for optimization},
  author={Wolpert, David H and Macready, William G},
  journal={IEEE transactions on evolutionary computation},
  volume={1},
  number={1},
  pages={67--82},
  year={2002},
  publisher={IEEE},
  url = {https://ieeexplore.ieee.org/abstract/document/585893}
}

@article{funsearch,
  author  = {Romera-Paredes, Bernardino and Barekatain, Mohammadamin and Novikov, Alexander and Balog, Matej and Kumar, M. Pawan and Dupont, Emilien and Ruiz, Francisco J. R. and Ellenberg, Jordan S. and Wang, Pengming and Fawzi, Omar and Kohli, Pushmeet and Fawzi, Alhussein},
  title   = {Mathematical discoveries from program search with large language models},
  journal = {Nature},
  year    = {2024},
  volume  = {625},
  number  = {7995},
  pages   = {468--475},
  doi     = {10.1038/s41586-023-06924-6},
  url     = {https://doi.org/10.1038/s41586-023-06924-6},
  issn    = {1476-4687}
}

@inproceedings{alphaevolve,
author = {Cui, Can and Wang, Wei and Zhang, Meihui and Chen, Gang and Luo, Zhaojing and Ooi, Beng Chin},
title = {AlphaEvolve: A Learning Framework to Discover Novel Alphas in Quantitative Investment},
year = {2021},
isbn = {9781450383431},
publisher = {Association for Computing Machinery},
address = {New York, NY, USA},
url = {https://doi.org/10.1145/3448016.3457324},
doi = {10.1145/3448016.3457324},
booktitle = {Proceedings of the 2021 International Conference on Management of Data},
pages = {2208–2216},
numpages = {9},
location = {Virtual Event, China},
series = {SIGMOD '21}
}

@misc{huggingface2024aime,
  author       = {{Hugging Face H4}},
  title        = {AIME 2024 Dataset},
  year         = {2024},
  howpublished = {\url{https://huggingface.co/datasets/HuggingFaceH4/aime_2024}},
  note         = {Accessed: 2025-05-16}
}

@inproceedings{yao2023reactsynergizingreasoningacting,
      title={ReAct: Synergizing Reasoning and Acting in Language Models},
      author={Shunyu Yao and Jeffrey Zhao and Dian Yu and Nan Du and Izhak Shafran and Karthik Narasimhan and Yuan Cao},
      booktitle={Proceedings of the Eleventh International Conference on Learning Representations},
      year={2023},
      url={https://openreview.net/forum?id=WE_vluYUL-X},
}

@misc{wang2023planandsolvepromptingimprovingzeroshot,
      title={Plan-and-Solve Prompting: Improving Zero-Shot Chain-of-Thought Reasoning by Large Language Models}, 
      author={Lei Wang and Wanyu Xu and Yihuai Lan and Zhiqiang Hu and Yunshi Lan and Roy Ka-Wei Lee and Ee-Peng Lim},
      year={2023},
      eprint={2305.04091},
      archivePrefix={arXiv},
      primaryClass={cs.CL},
      url={https://arxiv.org/abs/2305.04091}, 
}

@misc{wang2023voyageropenendedembodiedagent,
      title={Voyager: An Open-Ended Embodied Agent with Large Language Models}, 
      author={Guanzhi Wang and Yuqi Xie and Yunfan Jiang and Ajay Mandlekar and Chaowei Xiao and Yuke Zhu and Linxi Fan and Anima Anandkumar},
      year={2023},
      eprint={2305.16291},
      archivePrefix={arXiv},
      primaryClass={cs.AI},
      url={https://arxiv.org/abs/2305.16291}, 
}

@article{Besta_Blach_Kubicek_Gerstenberger_Podstawski_Gianinazzi_Gajda_Lehmann_Niewiadomski_Nyczyk_Hoefler_2024, title={Graph of Thoughts: Solving Elaborate Problems with Large Language Models}, volume={38}, url={https://ojs.aaai.org/index.php/AAAI/article/view/29720}, DOI={10.1609/aaai.v38i16.29720}, abstractNote={We introduce Graph of Thoughts (GoT): a framework that
advances prompting capabilities in large language models
(LLMs) beyond those offered by paradigms such as Chain-of-Thought or Tree of Thoughts (ToT). The key idea and primary advantage of GoT is the ability to model the information generated by an LLM as an arbitrary graph, where units of information (&quot;LLM thoughts&quot;) are vertices, and edges correspond
to dependencies between these vertices. This approach enables combining arbitrary LLM thoughts into synergistic outcomes, distilling the essence of whole networks of thoughts,
or enhancing thoughts using feedback loops. We illustrate
that GoT offers advantages over state of the art on different
tasks, for example increasing the quality of sorting by 62%
over ToT, while simultaneously reducing costs by &gt;31%.
We ensure that GoT is extensible with new thought transformations and thus can be used to spearhead new prompting
schemes. This work brings the LLM reasoning closer to human thinking or brain mechanisms such as recurrence, both
of which form complex networks}, number={16}, journal={Proceedings of the AAAI Conference on Artificial Intelligence}, author = {Besta, Maciej and Blach, Nils and Kubicek, Ales and Gerstenberger, Robert and Podstawski, Micha\l{} and Gianinazzi, Lukas and Gajda, Joanna and Lehmann, Tomasz and Niewiadomski, Hubert and Nyczyk, Piotr and Hoefler, Torsten}, year={2024}, month={Mar.}, pages={17682-17690} }

@misc{chen2023programthoughtspromptingdisentangling,
      title={Program of Thoughts Prompting: Disentangling Computation from Reasoning for Numerical Reasoning Tasks}, 
      author={Wenhu Chen and Xueguang Ma and Xinyi Wang and William W. Cohen},
      year={2023},
      eprint={2211.12588},
      archivePrefix={arXiv},
      primaryClass={cs.CL},
      url={https://arxiv.org/abs/2211.12588}, 
}

@InProceedings{pmlr-v202-gao23f,
  title = 	 {{PAL}: Program-aided Language Models},
  author =       {Gao, Luyu and Madaan, Aman and Zhou, Shuyan and Alon, Uri and Liu, Pengfei and Yang, Yiming and Callan, Jamie and Neubig, Graham},
  booktitle = 	 {Proceedings of the 40th International Conference on Machine Learning},
  pages = 	 {10764--10799},
  year = 	 {2023},
  volume = 	 {202},
  series = 	 {Proceedings of Machine Learning Research},
  month = 	 {23--29 Jul},
  publisher =    {PMLR},
  url = 	 {https://proceedings.mlr.press/v202/gao23f.html}
}

@article{HOGG19961,
title = {Phase transitions and the search problem},
journal = {Artificial Intelligence},
volume = {81},
number = {1},
pages = {1-15},
year = {1996},
note = {Frontiers in Problem Solving: Phase Transitions and Complexity},
issn = {0004-3702},
doi = {https://doi.org/10.1016/0004-3702(95)00044-5},
url = {https://www.sciencedirect.com/science/article/pii/0004370295000445},
author = {Tad Hogg and Bernardo A. Huberman and Colin P. Williams}
}

@misc{Kaplan2020ScalingLaws,
      title={Scaling Laws for Neural Language Models}, 
      author={Jared Kaplan and Sam McCandlish and Tom Henighan and Tom B. Brown and Benjamin Chess and Rewon Child and Scott Gray and Alec Radford and Jeffrey Wu and Dario Amodei},
      year={2020},
      eprint={2001.08361},
      archivePrefix={arXiv},
      primaryClass={cs.LG},
      url={https://arxiv.org/abs/2001.08361}, 
}

@misc{bran2023chemcrow,
      title={ChemCrow: Augmenting large-language models with chemistry tools}, 
      author={Andres M Bran and Sam Cox and Oliver Schilter and Carlo Baldassari and Andrew D White and Philippe Schwaller},
      year={2023},
      eprint={2304.05376},
      archivePrefix={arXiv},
      primaryClass={physics.chem-ph},
      url={https://arxiv.org/abs/2304.05376}, 
}

@inproceedings{EoH,
      title={Evolution of Heuristics: Towards Efficient Automatic Algorithm Design Using Large Language Model},
      author={Fei Liu and Xialiang Tong and Mingxuan Yuan and Xi Lin and Fu Luo and Zhenkun Wang and Zhichao Lu and Qingfu Zhang},
      booktitle={Proceedings of the 41st International Conference on Machine Learning},
      series={PMLR},
      volume={235},
      pages={32201--32223},
      year={2024},
      url={https://proceedings.mlr.press/v235/liu24bs.html},
}

@inproceedings{Park2023,
author = {Park, Joon Sung and O'Brien, Joseph and Cai, Carrie Jun and Morris, Meredith Ringel and Liang, Percy and Bernstein, Michael S.},
title = {Generative Agents: Interactive Simulacra of Human Behavior},
year = {2023},
isbn = {9798400701320},
publisher = {Association for Computing Machinery},
address = {New York, NY, USA},
url = {https://doi.org/10.1145/3586183.3606763},
doi = {10.1145/3586183.3606763},
booktitle = {Proceedings of the 36th Annual ACM Symposium on User Interface Software and Technology},
articleno = {2},
numpages = {22},
location = {San Francisco, CA, USA},
series = {UIST '23}
}

@misc{qwen2025qwen25technicalreport,
      title={Qwen2.5 Technical Report}, 
      author={Qwen Team},
      year={2025},
      eprint={2412.15115},
      archivePrefix={arXiv},
      primaryClass={cs.CL},
      url={https://arxiv.org/abs/2412.15115}, 
}

@misc{kim2025sciencescalingagentsystems,
      title={Towards a Science of Scaling Agent Systems}, 
      author={Yubin Kim and Ken Gu and Chanwoo Park and Chunjong Park and Samuel Schmidgall and A. Ali Heydari and Yao Yan and Zhihan Zhang and Yuchen Zhuang and Mark Malhotra and Paul Pu Liang and Hae Won Park and Yuzhe Yang and Xuhai Xu and Yilun Du and Shwetak Patel and Tim Althoff and Daniel McDuff and Xin Liu},
      year={2025},
      eprint={2512.08296},
      archivePrefix={arXiv},
      primaryClass={cs.AI},
      url={https://arxiv.org/abs/2512.08296}, 
}

@article{wang2023surveyautonomousagents,
      title={A survey on large language model based autonomous agents},
      author={Lei Wang and Chen Ma and Xueyang Feng and Zeyu Zhang and Hao Yang and Jingsen Zhang and Zhiyuan Chen and Jiakai Tang and Xu Chen and Yankai Lin and Wayne Xin Zhao and Zhewei Wei and Ji-Rong Wen},
      journal={Frontiers of Computer Science},
      volume={18},
      number={6},
      pages={186345},
      year={2024},
      publisher={Springer},
      url={https://link.springer.com/article/10.1007/s11704-024-40231-1},
}

@article{xi2023risepotentialagents,
      title={The rise and potential of large language model based agents: A survey},
      author={Zhiheng Xi and Wenxiang Chen and Xin Guo and Wei He and Yiwen Ding and Boyang Hong and Ming Zhang and Junzhe Wang and Senjie Jin and Enyu Zhou and Rui Zheng and Xiaoran Fan and Xiao Wang and Limao Xiong and Yuhao Zhou and Weiran Wang and Changhao Jiang and Yicheng Zou and Xiangyang Liu and Zhangyue Yin and Shihan Dou and Rongxiang Weng and Wensen Cheng and Qi Zhang and Wenjuan Qin and Yongyan Zheng and Xipeng Qiu and Xuanjing Huang and Tao Gui},
      journal={Science China Information Sciences},
      volume={68},
      pages={121101},
      year={2025},
      publisher={Springer},
      url={https://link.springer.com/article/10.1007/s11432-024-4222-0},
}

@inproceedings{schick2023toolformer,
      title={Toolformer: Language Models Can Teach Themselves to Use Tools},
      author={Timo Schick and Jane Dwivedi-Yu and Roberto Dess{\`i} and Roberta Raileanu and Maria Lomeli and Luke Zettlemoyer and Nicola Cancedda and Thomas Scialom},
      booktitle={Advances in Neural Information Processing Systems},
      volume={36},
      year={2023},
      url={https://proceedings.neurips.cc/paper_files/paper/2023/hash/d842425e4bf79ba039352da0f658a906-Abstract-Conference.html},
}

@inproceedings{madaan2023selfrefine,
      title={Self-Refine: Iterative Refinement with Self-Feedback},
      author={Aman Madaan and Niket Tandon and Prakhar Gupta and Skyler Hallinan and Luyu Gao and Sarah Wiegreffe and Uri Alon and Nouha Dziri and Shrimai Prabhumoye and Yiming Yang and Shashank Gupta and Bodhisattwa Prasad Majumder and Katherine Hermann and Sean Welleck and Amir Yazdanbakhsh and Peter Clark},
      booktitle={Advances in Neural Information Processing Systems},
      volume={36},
      year={2023},
      url={https://proceedings.neurips.cc/paper_files/paper/2023/hash/91edff07232fb1b55a505a9e9f6c0ff3-Abstract-Conference.html},
}

@inproceedings{shinn2023reflexion,
      title={Reflexion: Language Agents with Verbal Reinforcement Learning},
      author={Noah Shinn and Federico Cassano and Ashwin Gopinath and Karthik Narasimhan and Shunyu Yao},
      booktitle={Advances in Neural Information Processing Systems},
      volume={36},
      year={2023},
      url={https://proceedings.neurips.cc/paper_files/paper/2023/hash/1b44b878bb782e6954cd888628510e90-Abstract-Conference.html},
}

@inproceedings{wei2022chainofthought,
      title={Chain-of-Thought Prompting Elicits Reasoning in Large Language Models},
      author={Jason Wei and Xuezhi Wang and Dale Schuurmans and Maarten Bosma and Brian Ichter and Fei Xia and Ed Chi and Quoc Le and Denny Zhou},
      booktitle={Advances in Neural Information Processing Systems},
      volume={35},
      year={2022},
      url={https://proceedings.neurips.cc/paper_files/paper/2022/hash/9d5609613524ecf4f15af0f7b31abca4-Abstract-Conference.html},
}

@inproceedings{wang2022selfconsistency,
      title={Self-Consistency Improves Chain of Thought Reasoning in Language Models},
      author={Xuezhi Wang and Jason Wei and Dale Schuurmans and Quoc Le and Ed Chi and Sharan Narang and Aakanksha Chowdhery and Denny Zhou},
      booktitle={Proceedings of the Eleventh International Conference on Learning Representations},
      year={2023},
      url={https://openreview.net/forum?id=1PL1NIMMrw},
}

@inproceedings{ouyang2022traininglanguagemodels,
      title={Training language models to follow instructions with human feedback},
      author={Long Ouyang and Jeff Wu and Xu Jiang and Diogo Almeida and Carroll L. Wainwright and Pamela Mishkin and Chong Zhang and Sandhini Agarwal and Katarina Slama and Alex Ray and John Schulman and Jacob Hilton and Fraser Kelton and Luke Miller and Maddie Simens and Amanda Askell and Peter Welinder and Paul Christiano and Jan Leike and Ryan Lowe},
      booktitle={Advances in Neural Information Processing Systems},
      volume={35},
      year={2022},
      url={https://proceedings.neurips.cc/paper_files/paper/2022/hash/b1efde53be364a73914f58805a001731-Abstract-Conference.html},
}

@inproceedings{yao2023treeofthoughts,
      title={Tree of Thoughts: Deliberate Problem Solving with Large Language Models},
      author={Shunyu Yao and Dian Yu and Jeffrey Zhao and Izhak Shafran and Thomas L. Griffiths and Yuan Cao and Karthik Narasimhan},
      booktitle={Advances in Neural Information Processing Systems},
      volume={36},
      year={2023},
      url={https://proceedings.neurips.cc/paper_files/paper/2023/hash/271db9922b8d1f4dd7aaef84ed5ac703-Abstract-Conference.html},
}

@book{de2017linear,
  title={Linear response theory: an analytic-algebraic approach},
  author={De Nittis, Giuseppe and Lein, Max},
  year={2017},
  publisher={Springer}
}

@misc{yang2025qwen3technicalreport,
      title={Qwen3 Technical Report}, 
      author={An Yang and Anfeng Li and Baosong Yang and Beichen Zhang and Binyuan Hui and Bo Zheng and Bowen Yu and Chang Gao and Chengen Huang and Chenxu Lv and Chujie Zheng and Dayiheng Liu and Fan Zhou and Fei Huang and Feng Hu and Hao Ge and Haoran Wei and Huan Lin and Jialong Tang and Jian Yang and Jianhong Tu and Jianwei Zhang and Jianxin Yang and Jiaxi Yang and Jing Zhou and Jingren Zhou and Junyang Lin and Kai Dang and Keqin Bao and Kexin Yang and Le Yu and Lianghao Deng and Mei Li and Mingfeng Xue and Mingze Li and Pei Zhang and Peng Wang and Qin Zhu and Rui Men and Ruize Gao and Shixuan Liu and Shuang Luo and Tianhao Li and Tianyi Tang and Wenbiao Yin and Xingzhang Ren and Xinyu Wang and Xinyu Zhang and Xuancheng Ren and Yang Fan and Yang Su and Yichang Zhang and Yinger Zhang and Yu Wan and Yuqiong Liu and Zekun Wang and Zeyu Cui and Zhenru Zhang and Zhipeng Zhou and Zihan Qiu},
      year={2025},
      eprint={2505.09388},
      archivePrefix={arXiv},
      primaryClass={cs.CL},
      url={https://arxiv.org/abs/2505.09388}, 
}

@misc{durante2024agentaisurveyingfoundations,
      title={Agent AI: Surveying the Horizons of Multimodal Interaction}, 
      author={Zane Durante and Qiuyuan Huang and Naoki Wake and Ran Gong and Jae Sung Park and Bidipta Sarkar and Rohan Taori and Yusuke Noda and Demetri Terzopoulos and Yejin Choi and Katsushi Ikeuchi and Hoi Vo and Li Fei-Fei and Jianfeng Gao},
      year={2024},
      eprint={2401.03568},
      archivePrefix={arXiv},
      primaryClass={cs.AI},
      url={https://arxiv.org/abs/2401.03568}, 
}

@misc{hu2024automateddesignagenticsystems,
      title={Automated Design of Agentic Systems}, 
      author={Shengran Hu and Cong Lu and Jeff Clune},
      year={2025},
      eprint={2408.08435},
      archivePrefix={arXiv},
      primaryClass={cs.AI},
      url={https://arxiv.org/abs/2408.08435}, 
}

@book{cover2006elements,
  title={Elements of Information Theory},
  author={Cover, Thomas M. and Thomas, Joy A.},
  edition={2},
  year={2006},
  publisher={Wiley-Interscience},
  address={Hoboken, NJ},
  isbn={978-0-471-24195-9},
  url={https://onlinelibrary.wiley.com/doi/book/10.1002/047174882X}
}

@article{shannon1948mathematical,
  title={A mathematical theory of communication},
  author={Shannon, Claude E.},
  journal={The Bell System Technical Journal},
  volume={27},
  number={3},
  pages={379--423},
  year={1948},
  publisher={Nokia Bell Labs},
  doi={10.1002/j.1538-7305.1948.tb01338.x},
  url={https://ieeexplore.ieee.org/document/6773024}
}

@book{kellerer2004knapsack,
  title={Knapsack Problems},
  author={Kellerer, Hans and Pferschy, Ulrich and Pisinger, David},
  year={2004},
  publisher={Springer},
  address={Berlin},
  isbn={978-3-540-40286-2},
  doi={10.1007/978-3-540-24777-7},
  url={https://link.springer.com/book/10.1007/978-3-540-24777-7}
}

@book{russell2021artificial,
  title={Artificial Intelligence: A Modern Approach},
  author={Russell, Stuart and Norvig, Peter},
  edition={4},
  year={2021},
  publisher={Pearson},
  address={Hoboken, NJ},
  isbn={978-0-13-461099-3},
  url={https://www.pearson.com/en-us/subject-catalog/p/artificial-intelligence-a-modern-approach/P200000003500/9780137505135}
}

@article{kubo1957statistical,
  title={Statistical-mechanical theory of irreversible processes. {I}. {General} theory and simple applications to magnetic and conduction problems},
  author={Kubo, Ryogo},
  journal={Journal of the Physical Society of Japan},
  volume={12},
  number={6},
  pages={570--586},
  year={1957},
  publisher={The Physical Society of Japan},
  doi={10.1143/JPSJ.12.570},
  url={https://doi.org/10.1143/JPSJ.12.570}
}

@incollection{good1966speculations,
  title={Speculations concerning the first ultraintelligent machine},
  author={Good, Irving John},
  booktitle={Advances in Computers},
  volume={6},
  pages={31--88},
  year={1966},
  publisher={Academic Press},
  doi={10.1016/S0065-2458(08)60418-0},
  url={https://doi.org/10.1016/S0065-2458(08)60418-0}
}

@article{lechatelier1884,
  title={Sur un \'enonc\'e g\'en\'eral des lois des \'equilibres chimiques},
  author={Le Chatelier, Henry Louis},
  journal={Comptes rendus de l'Acad\'emie des sciences},
  volume={99},
  pages={786--789},
  year={1884},
  url={https://gallica.bnf.fr/ark:/12148/bpt6k3055h/f786.item}
}

@book{condorcet1785essai,
  title={Essai sur l'application de l'analyse \`a la probabilit\'e des d\'ecisions rendues \`a la pluralit\'e des voix},
  author={de Condorcet, Marie Jean Antoine Nicolas de Caritat},
  year={1785},
  publisher={Imprimerie Royale, Paris},
  note={Reprinted by Chelsea, New York, 1972},
  url={https://gallica.bnf.fr/ark:/12148/bpt6k417181}
}

@inproceedings{hoffmann2022training,
  title={Training Compute-Optimal Large Language Models},
  author={Hoffmann, Jordan and Borgeaud, Sebastian and Mensch, Arthur and Buchatskaya, Elena and Cai, Trevor and Rutherford, Eliza and de Las Casas, Diego and Hendricks, Lisa Anne and Welbl, Johannes and Clark, Aidan and Hennigan, Tom and Noland, Eric and Millican, Katie and van den Driessche, George and Damoc, Bogdan and Guy, Aurelia and Osindero, Simon and Simonyan, Karen and Elsen, Erich and Rae, Jack W. and Vinyals, Oriol and Sifre, Laurent},
  booktitle={Advances in Neural Information Processing Systems},
  volume={35},
  year={2022},
  url={https://arxiv.org/abs/2203.15556}
}

@misc{snell2024scaling,
      title={Scaling LLM Test-Time Compute Optimally can be More Effective than Scaling Model Parameters}, 
      author={Charlie Snell and Jaehoon Lee and Kelvin Xu and Aviral Kumar},
      year={2024},
      eprint={2408.03314},
      archivePrefix={arXiv},
      primaryClass={cs.LG},
      url={https://arxiv.org/abs/2408.03314}, 
}

@article{li2022alphacode,
  title={Competition-Level Code Generation with {AlphaCode}},
  author={Li, Yujia and Choi, David and Chung, Junyoung and Kushman, Nate and Schrittwieser, Julian and Leblond, R\'{e}mi and Eccles, Tom and Keeling, James and Gimeno, Felix and Dal Lago, Agustin and Hubert, Thomas and Choy, Peter and de Masson d'Autume, Cyprien and Babuschkin, Igor and Chen, Xinyun and Huang, Po-Sen and Welbl, Johannes and Gowal, Sven and Cherepanov, Alexey and Molloy, James and Mankowitz, Daniel J. and Sutherland Robson, Esme and Kohli, Pushmeet and de Freitas, Nando and Kavukcuoglu, Koray and Vinyals, Oriol},
  journal={Science},
  volume={378},
  number={6624},
  pages={1092--1097},
  year={2022},
  doi={10.1126/science.abq1158},
  url={https://arxiv.org/abs/2203.07814}
}

@misc{cobbe2021training,
  title={Training Verifiers to Solve Math Word Problems},
  author={Karl Cobbe and Vineet Kosaraju and Mohammad Bavarian and Mark Chen and Heewoo Jun and Lukasz Kaiser and Matthias Plappert and Jerry Tworek and Jacob Hilton and Reiichiro Nakano and Christopher Hesse and John Schulman},
  year={2021},
  eprint={2110.14168},
  archivePrefix={arXiv},
  primaryClass={cs.LG},
  url={https://arxiv.org/abs/2110.14168}
}

@misc{brown2024large,
      title={Large Language Monkeys: Scaling Inference Compute with Repeated Sampling}, 
      author={Bradley Brown and Jordan Juravsky and Ryan Ehrlich and Ronald Clark and Quoc V. Le and Christopher Ré and Azalia Mirhoseini},
      year={2024},
      eprint={2407.21787},
      archivePrefix={arXiv},
      primaryClass={cs.LG},
      url={https://arxiv.org/abs/2407.21787}, 
}

@inproceedings{zelikman2022star,
  title={{STaR}: Bootstrapping Reasoning With Reasoning},
  author={Eric Zelikman and Yuhuai Wu and Jesse Mu and Noah D. Goodman},
  booktitle={Advances in Neural Information Processing Systems},
  volume={35},
  year={2022},
  url={https://arxiv.org/abs/2203.14465}
}

@inproceedings{singh2024beyond,
  title={Beyond Human Data: Scaling Self-Training for Problem-Solving with Language Models},
  author={Singh, Avi and Co-Reyes, John D. and Agarwal, Rishabh and Anand, Ankesh and Patil, Piyush and Liu, Peter J. and Harrison, James and Lee, Jaehoon and Xu, Kelvin and Parisi, Aaron and Kumar, Abhishek and Alemi, Alex and Rizkowsky, Alex and Nova, Azade and Adlam, Ben and Bohnet, Bernd and Elsayed, Gamaleldin and Sedghi, Hanie and Mordatch, Igor and Simpson, Isabelle and Gur, Izzeddin and Snoek, Jasper and Pennington, Jeffrey and Hron, Jiri and Kenealy, Kathleen and Swersky, Kevin and Mahajan, Kshiteej and Houlsby, Neil and Dyer, Ethan and Tran, Dustin and Zhou, Denny and Petrov, Slav and Gulcehre, Caglar},
  booktitle={Advances in Neural Information Processing Systems},
  volume={37},
  year={2024},
  url={https://arxiv.org/abs/2312.06585}
}

@misc{hosseini2024vstar,
      title={V-STaR: Training Verifiers for Self-Taught Reasoners}, 
      author={Arian Hosseini and Xingdi Yuan and Nikolay Malkin and Aaron Courville and Alessandro Sordoni and Rishabh Agarwal},
      year={2024},
      eprint={2402.06457},
      archivePrefix={arXiv},
      primaryClass={cs.LG},
      url={https://arxiv.org/abs/2402.06457}, 
}

@misc{openai2024o1systemcard,
  title={{OpenAI} o1 System Card},
  author={{OpenAI}},
  year={2024},
  eprint={2412.16720},
  archivePrefix={arXiv},
  primaryClass={cs.AI},
  url={https://arxiv.org/abs/2412.16720}
}

@misc{deepseek2025r1,
  title={{DeepSeek-R1}: Incentivizing Reasoning Capability in {LLMs} via Reinforcement Learning},
  author={{DeepSeek-AI}},
  year={2025},
  eprint={2501.12948},
  archivePrefix={arXiv},
  primaryClass={cs.AI},
  url={https://arxiv.org/abs/2501.12948}
}

@misc{he2026urlvr,
      title={How Far Can Unsupervised RLVR Scale LLM Training?}, 
      author={Bingxiang He and Yuxin Zuo and Zeyuan Liu and Shangziqi Zhao and Zixuan Fu and Junlin Yang and Cheng Qian and Kaiyan Zhang and Yuchen Fan and Ganqu Cui and Xiusi Chen and Youbang Sun and Xingtai Lv and Xuekai Zhu and Li Sheng and Ran Li and Huan-ang Gao and Yuchen Zhang and Bowen Zhou and Zhiyuan Liu and Ning Ding},
      year={2026},
      eprint={2603.08660},
      archivePrefix={arXiv},
      primaryClass={cs.LG},
      url={https://arxiv.org/abs/2603.08660}, 
}

% ============================================================
%  EXTENDED DATA
% ============================================================
\clearpage
\section*{Extended Data}

\newcounter{edfigure}
\renewcommand{\theedfigure}{\arabic{edfigure}}

\begin{figure*}[!t]
    \centering
    \includegraphics[width=\textwidth]{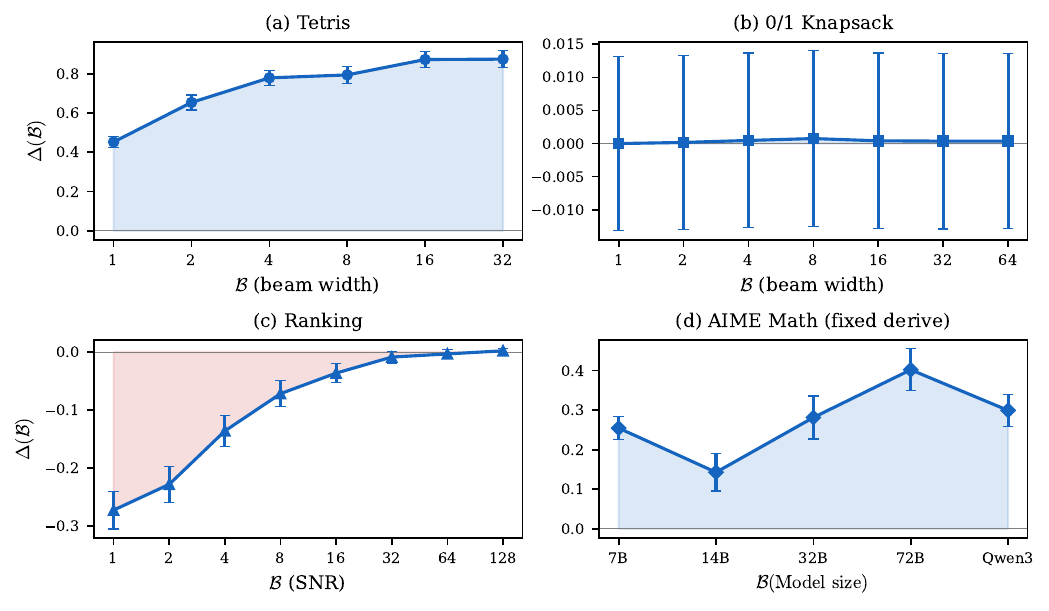}
    \refstepcounter{edfigure}
    \caption*{\textbf{Extended Data Fig. \theedfigure{} $|$ Averaged performance gap across domains.} The normalized performance gap $\Delta(\mathcal{B}) = \left(J(\mathcal{P}_\mathcal{B}) - J(\mathcal{P}'_\mathcal{B})\right)/\overline{J(\mathcal{P}_\mathcal{B})}$, averaged over all five LLMs, as a function of computational budget $\mathcal{B}$ for four domains. Blue shading indicates the regime where the base algorithm outperforms the LLM-derived strategy ($\Delta > 0$); red shading indicates the opposite. In Tetris, $\Delta$ grows monotonically. In Knapsack, $\Delta$ is negligible. In Ranking, $\Delta$ transitions from negative (LLM advantage at low SNR) to near zero. In AIME, $\Delta$ (averaged over $k \in \{15, 17, 19, 21\}$) remains positive across model sizes.}
    \label{fig:gap}
\end{figure*}

\begin{figure*}[!t]
    \centering
    \includegraphics[width=\textwidth]{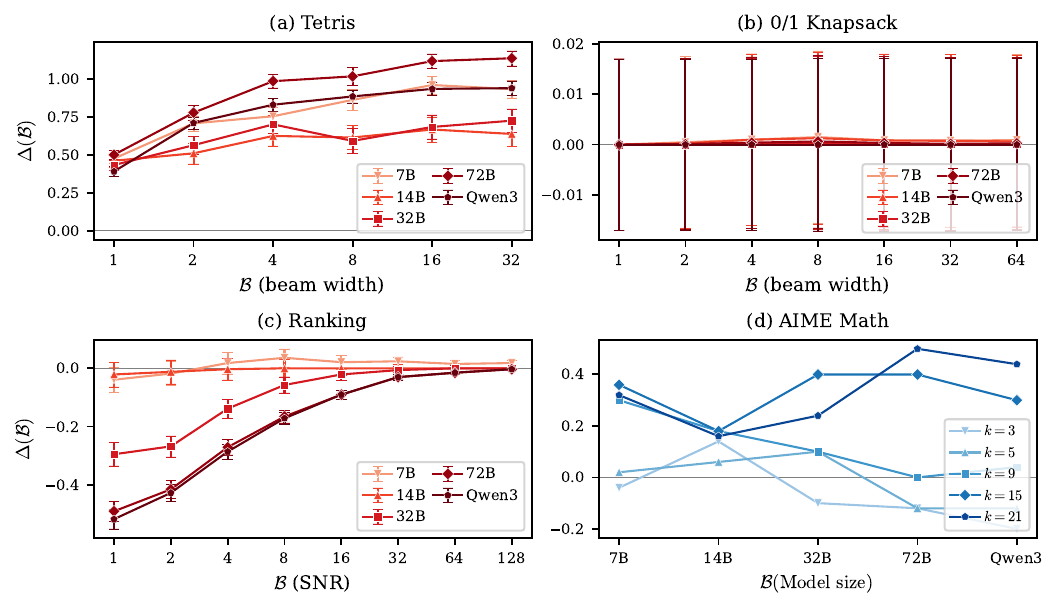}
    \refstepcounter{edfigure}
    \caption*{\textbf{Extended Data Fig. \theedfigure{} $|$ Per-model performance gap across domains.} The normalized performance gap $\Delta(\mathcal{B})$ broken down by individual model size (7B through Qwen3-Max) for each domain. In Tetris, 72B models show largest gaps. In Knapsack, all models produce negligible gaps. In Ranking, all models converge from negative to near-zero $\Delta$ as SNR increases. In AIME, the gap varies with both generator model size and number of samples $k$, with larger $k$ showing an increasing tendency with model size.}
    \label{fig:gap_models}
\end{figure*}

\begin{figure*}[!t]
    \centering
    \includegraphics[width=\textwidth]{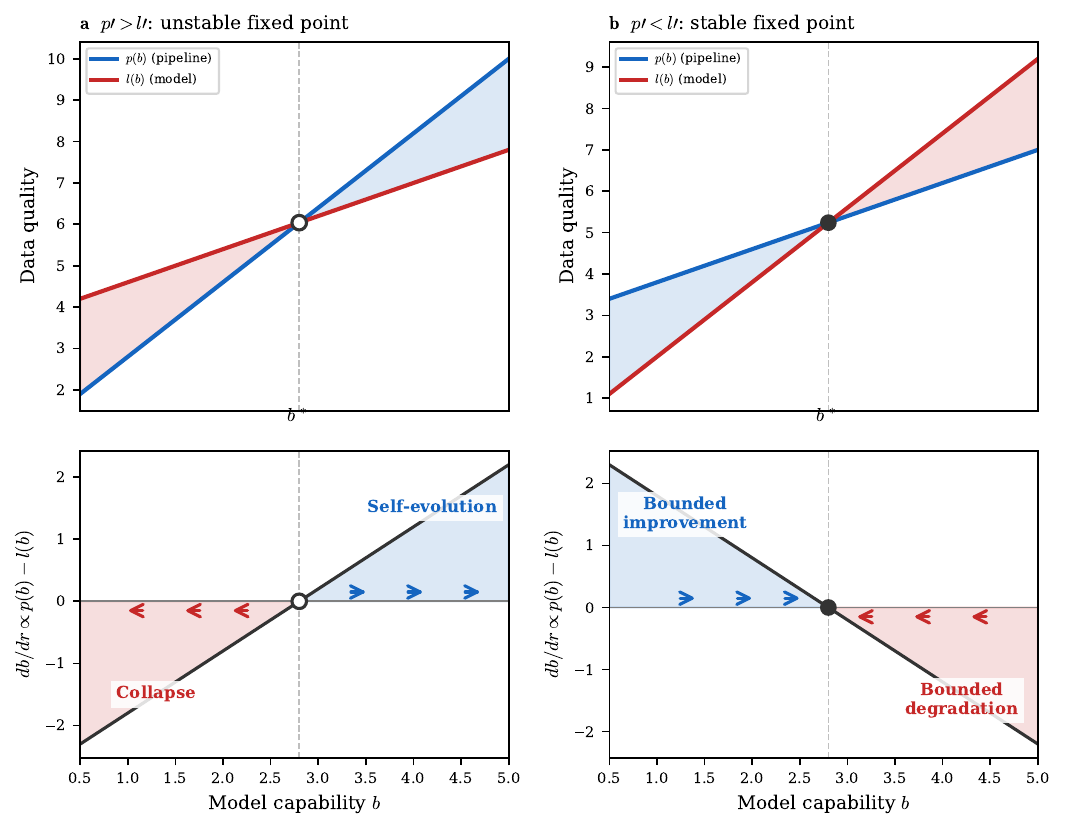}
    \refstepcounter{edfigure}
    \caption*{\textbf{Extended Data Fig. \theedfigure{} $|$ Illustration of the phenomenological theory of self-evolution dynamics.} Top row: data quality functions $p(b)$ (pipeline, blue) and $l(b)$ (model output, red) versus model capability $b$. Bottom row: phase portrait $db/dr \propto p(b) - l(b)$, with arrows indicating the flow direction. Open circle: unstable fixed point; filled circle: stable fixed point. \textbf{a}, $p'(b) > l'(b)$: the fixed point is a repeller, giving rise to a collapse phase ($b < b^*$) and a self-evolution phase ($b > b^*$). \textbf{b}, $p'(b) < l'(b)$: the fixed point is an attractor; improvement and degradation are both bounded. See Supplementary Note~1 for the full derivation.}
    \label{fig:selfevolution}
\end{figure*}

% ============================================================
%  SUPPLEMENTARY INFORMATION
% ============================================================
\clearpage
\setcounter{equation}{0}
\renewcommand{\theequation}{S\arabic{equation}}
\renewcommand{\theHequation}{S\arabic{equation}}
\setcounter{figure}{0}
\renewcommand{\thefigure}{S\arabic{figure}}

\section*{Supplementary Information}

\subsection*{Supplementary Note 1: Phenomenological theory of self-evolution}

The self-evolution argument in the main text can be formalized with a minimal dynamical model. Let $b$ denote the capability of a model, $p(b)$ the quality of training data produced by a data-generation pipeline constructed using a model of capability $b$, and $l(b)$ the quality of output generated directly by a model of capability $b$. When the model is trained on its own pipeline-generated data, the capability evolves according to
\begin{equation}
    \frac{db}{dr} = \eta\,[p(b) - l(b)],
    \label{eq:selfevolution}
\end{equation}
where $r$ is the cumulative training resource and $\eta > 0$ is a learning-rate constant. The driving term $p(b) - l(b)$ represents the gap between what the pipeline can produce and what the model currently outputs: when the pipeline generates higher-quality data than the model's own output ($p > l$), training improves the model; when the pipeline produces lower-quality data ($p < l$), training degrades it.

A fixed point $b^*$ satisfies $p(b^*) = l(b^*)$: the pipeline output quality matches the model's own output, so training produces no net change in capability. The stability of this fixed point is determined by the sign of $p'(b^*) - l'(b^*)$.

\textbf{Case 1: $p'(b) > l'(b)$ (repeller).} Since $p - l$ is an increasing function of $b$, the fixed point $b^*$ is unstable (Extended Data Fig.~\ref{fig:selfevolution}a). For $b < b^*$, $p(b) < l(b)$ and $db/dr < 0$: the pipeline produces data of lower quality than the model's own output, so training degrades capability, which further widens the gap (collapse phase with positive feedback). For $b > b^*$, $p(b) > l(b)$ and $db/dr > 0$: the pipeline data quality exceeds the model's output, so training continually improves the model and the improvement accelerates as the gap widens (self-evolution phase). The system thus exhibits a phase transition: whether the initial capability $b_0$ lies above or below the critical point $b^*$ determines whether the model undergoes unbounded self-evolution or irreversible collapse.

\textbf{Case 2: $p'(b) < l'(b)$ (attractor).} Since $p - l$ is a decreasing function of $b$, the fixed point $b^*$ is stable (Extended Data Fig.~\ref{fig:selfevolution}b). For $b < b^*$, $p(b) > l(b)$ and the model improves, but the improvement decelerates as $b$ approaches $b^*$ (bounded improvement). For $b > b^*$, $p(b) < l(b)$ and the model degrades, but the degradation likewise decelerates (bounded degradation). In both cases the system converges to $b^*$. There is no phase transition; training always produces a finite, bounded change in capability.

\textbf{Marginal case: $p'(b) = l'(b)$.} When the two slopes are equal, $p(b) - l(b)$ is a constant independent of $b$. If this constant is positive, the system is in a global self-evolution phase; if negative, it collapses globally. No fixed point exists and no phase transition occurs. In the linear model this case is degenerate, as it requires two parallel lines whose fate is determined entirely by the sign of the global offset.

\textbf{Connection to the susceptibility framework.} In the framework developed in the main text, the pipeline quality $p(b)$ corresponds to the effective performance of a nested architecture in which a model of capability $b$ serves as both generator and selector, while $l(b)$ corresponds to the performance of the base strategy (e.g., majority vote). The condition $p'(b) > l'(b)$ is then equivalent to the nested total sensitivity $\alpha_\mathrm{total} > 1$ (positive coupling regime, equation~(\ref{eq:nested}) in the main text), whereas $p'(b) < l'(b)$ corresponds to $\alpha_\mathrm{total} < 1$ (negative coupling or decoupled regime). Within the hypothesis framework of the main text, the requirement of nested co-scaling to realize $\alpha_\mathrm{total} > 1$ can thus be restated dynamically: self-evolution is possible only when the pipeline's data quality responds to model capability faster than the model's own output quality does.

\end{document}